\documentclass[11pt,a4paper]{article}
\usepackage[hyperref]{emnlp2020}
\usepackage{times}
\usepackage{latexsym}

\usepackage{setspace}

\usepackage{microtype}

\aclfinalcopy %
\def\aclpaperid{168} %

\usepackage{mathtools}
\usepackage{amsfonts} 
\usepackage{amsmath}
\usepackage{subcaption}

\usepackage{times}
\usepackage{latexsym}
\usepackage{amsmath}
\usepackage{graphicx}
\usepackage{xcolor}
\usepackage{caption}
\usepackage{xspace}
\usepackage{booktabs}
\usepackage{float}
\usepackage{array}
\usepackage{colortbl}
\usepackage{soul}
\usepackage{ascii}
\usepackage{bbm}

\usepackage{CJKutf8}

\usepackage{hyperref}
\hypersetup{
  	colorlinks=true,
  	citecolor=blue,
    linkcolor=black,
    filecolor=black,
    urlcolor=black}

\author{Liye Fu \\
	Cornell University \\
  \texttt{liye@cs.cornell.edu} \\\And
	Susan R. Fussell \\
	Cornell University \\
	\texttt{sfussell@cornell.edu} \\\And
	Cristian Danescu-Niculescu-Mizil \\
 	Cornell University \\
  \texttt{\hspace{0.3 in} cristian@cs.cornell.edu} \\}

\definecolor{lightergray}{RGB}{240, 240, 240}
\definecolor{lightblue}{rgb}{0.7,0.8,1.0}
\definecolor{lightyellow}{rgb}{1.00,0.93,0.55}
\definecolor{lightpink}{rgb}{1.0,0.8,0.8}
\definecolor{darkgreen}{rgb}{0.24,0.57,0.25}

\newif\ifshowcomments
\showcommentstrue

\usepackage{xcolor}

\newcommand{\xhdr}[1]{{\noindent\bfseries #1.}}

\newcommand{\drg}{\textsc{drg}\xspace}

\newcommand{\gst}{\texttt{G-GST}\xspace}
\newcommand{\ilp}{\textsc{ilp}\xspace}
\newcommand{\attr}{markers\xspace}
\newcommand{\mae}{\normalsize\textsc{mae}\xspace}

\DeclareTextFontCommand{\textascii}{\asciifamily}
\newcommand{\speaker}{\textascii{sender}\xspace}
\newcommand{\listener}{\textascii{receiver}\xspace}
\newcommand{\sender}{{\speaker}\xspace}
\newcommand{\channel}{\textascii{channel}\xspace}
\newcommand{\receiver}{{\listener}\xspace}

\newcommand{\senders}{\textascii{senders}\xspace}
\newcommand{\channels}{\textascii{channels}\xspace}
\newcommand{\receivers}{\textascii{receivers}\xspace}

\newcommand{\context}{circumstance\xspace}
\newcommand{\contexts}{circumstances\xspace}
\newcommand{\content}{context\xspace}

\newcommand{\interpretation}{perception\xspace}
\newcommand{\degree}{level\xspace}
\newcommand{\preserving}{intention-preserving\xspace}

\DeclareMathOperator*{\argmin}{arg\,min}

\newcommand{\cfunc}{$f_{c}$\xspace}
\newcommand{\fsend}{$f_{send}$\xspace}
\newcommand{\frec}{$f_{rec}$\xspace}

\newcommand{\strset}{\mathcal{S}}
\newcommand{\strin}{\mathcal{S}_{in}}
\newcommand{\strout}{\mathcal{S}_{out}}

\newcommand{\indicator}{\mathbbm{1}}

\newcommand{\could}{\textsf{\small Subjunctive}\xspace}
\newcommand{\pls}{\textsf{\small Please}\xspace}
\newcommand{\plsstart}{\textsf{\small Please.Start}\xspace}
\newcommand{\filler}{\textsf{\small Filler}\xspace}
\newcommand{\swearing}{\textsf{\small Swearing}\xspace}
\newcommand{\can}{\textsf{\small Indicative}\xspace}
\newcommand{\hedges}{\textsf{\small Hedges}\xspace}
\newcommand{\forme}{\textsf{\small For.Me}\xspace}
\newcommand{\foryou}{\textsf{\small For.You}\xspace}
\newcommand{\thx}{\textsf{\small Gratitude}\xspace}
\newcommand{\hello}{\textsf{\small Greeting}\xspace}
\newcommand{\apology}{\textsf{\small Apology}\xspace}
\newcommand{\affirmation}{\textsf{\small Affirmation}\xspace}
\newcommand{\just}{\textsf{\small Adverb.Just}\xspace}
\newcommand{\conj}{\textsf{\small Conj.Start}\xspace}
\newcommand{\actually}{\textsf{\small Actually}\xspace}
\newcommand{\btw}{\textsf{\small By.The.Way}\xspace}

\newcommand{\reassurance}{\textsf{\small Reassurance}\xspace}

\title{Facilitating the Communication of Politeness through \\ Fine-Grained Paraphrasing}

\begin{document}

\maketitle
\begin{abstract}

Aided by technology, people are increasingly able to communicate across geographical, cultural, and language barriers.
This ability also results in new challenges, as interlocutors need to adapt their communication approaches to increasingly diverse \context{s}. 
In this work, we take the first steps towards automatically assisting people in adjusting their language to a specific communication \context. 

As a case study, we focus on facilitating the accurate transmission of pragmatic intentions and introduce a methodology for suggesting paraphrases that achieve the intended level of politeness under a given communication circumstance.
We demonstrate the feasibility of this approach by evaluating our method in two realistic communication scenarios and show that it can reduce the potential for misalignment between the speaker's intentions and the listener's perceptions in both cases.

\end{abstract}

\section{Introduction}
\label{sec:intro}

Technological developments have greatly enhanced our communication experience, providing 
 the opportunity to overcome geographic, 
cultural
 and language barriers to interact with people from different backgrounds in diverse settings \cite{herring_computer-mediated_1996}. 
However, this opportunity brings additional challenges for the interlocutors, as they need to adjust their language to increasingly diverse communication \contexts.

As humans, we often make conscious attempts to account for the communication setting. 
For instance, we may simplify our expressions if we know our listener has relatively limited language proficiency, and we tend to {be more polite towards people with higher status}. 
However, managing these {stylistic}  adjustments can be cognitively taxing, especially when we are missing relevant information---e.g., the language proficiency or the status of a conversational partner we meet online.

\begin{figure}
 \centering
	\includegraphics[width=\columnwidth]{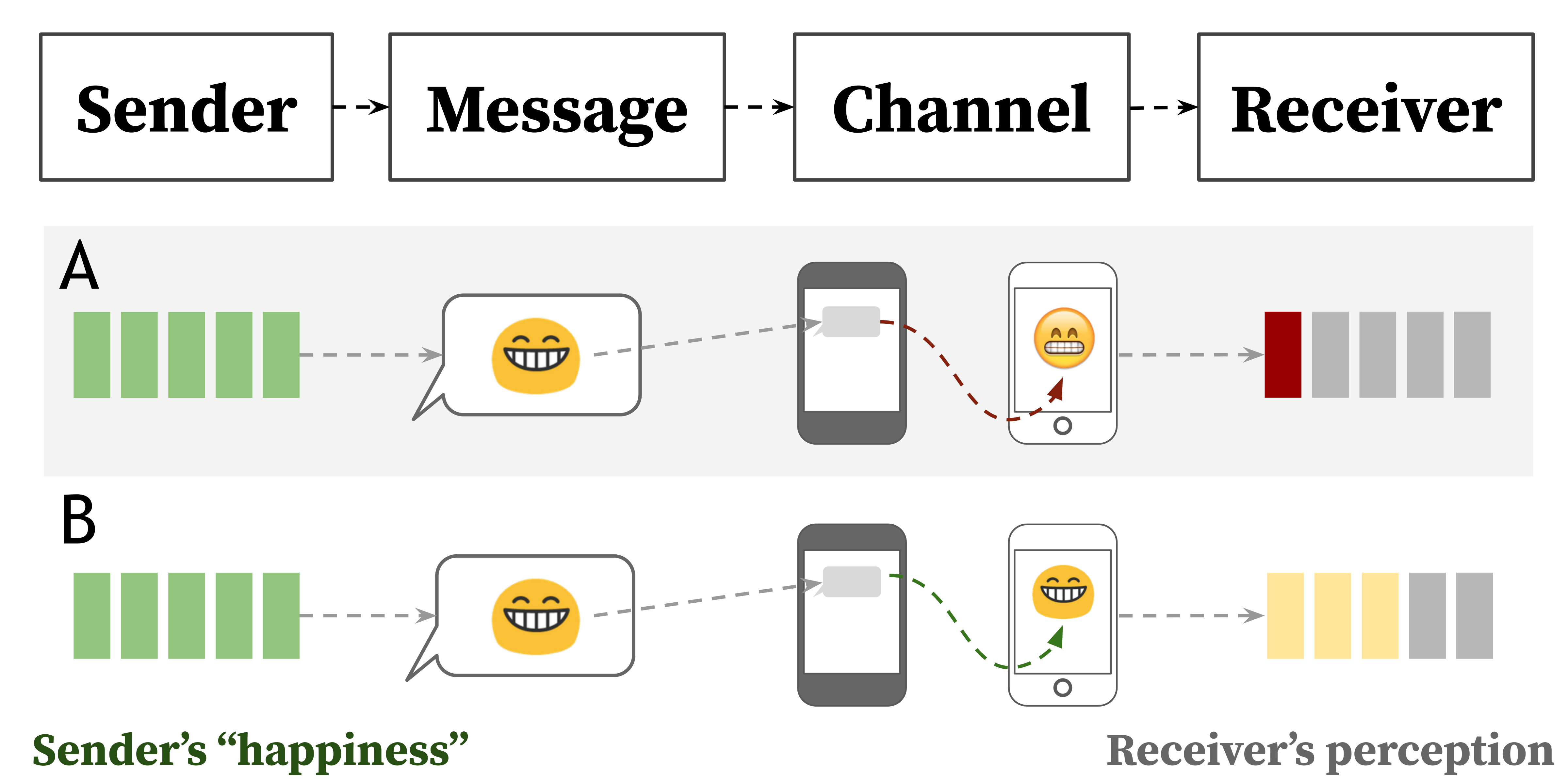}
	\caption{Berlo's Sender-Message-Channel-Receiver model suggests that the intended and perceived style of a message can be misaligned if: \textbf{A}. the \channel does not faithfully transmit the message, or \textbf{B}. the \receiver has a different reading of the message compared to the \sender. Examples are inspired by \citet{miller_blissfully_2016}.}

 \label{fig:smcr}
\end{figure}

If we do not adjust our language, we risk not properly conveying our pragmatic intentions \cite{thomas_cross-cultural_1983}.
In particular, Berlo's Sender-Message-Channel-Receiver model \cite{berlo_process_1960} points to two potential circumstance-specific causes for misalignments between intentions and perceptions (Figure~\ref{fig:smcr}). 
In this work we explore a method for assisting speakers to avoid such misalignments by suggesting for each message a paraphrase that is more likely to convey the original pragmatic intention when communicated in a given \context, as determined by the properties of the \sender, \channel, and \receiver.

As a case study, in this work, we focus on 
one particular pragmatic aspect: politeness.  
It is important to assist people to accurately transmit their intended politeness, as this interpersonal style \cite{biber_variation_1988} plays a key role in social interactions \cite{burke_mind_2008, murphy_impoliteness_2014,hu_read_2019, maaravi_and_2019}. 
Furthermore, politeness is known to be a \context-sensitive phenomenon \cite{kasper_cross-cultural_1990,herring_politeness_1994,forgas_asking_1998,mousavi_contrastive_2013}, making it a good case for our study. 
Concretely, we propose the task of generating a paraphrase for a given message that is more likely to 
{deliver} the intended level of politeness {after transmission} {(henceforth {\it \preserving})}, considering the properties of the \sender, \channel, and \receiver (Section \ref{sec:task}).

Taking the properties of the \channel into 
account is important because communication \channels may not always faithfully deliver messages (Figure~\ref{fig:smcr}A). 
For example, in translated communication, politeness signals can often be lost or corrupted \cite{allison_missed_2020}.
To demonstrate the potential of our framework in mitigating \channel-induced misunderstandings, we apply it to suggest paraphrases that are {\it safer to transmit}---i.e., less likely to have their politeness  
altered---over a commercial 
 machine translation 
  service.

We also need to account for the fact that the \sender and 
 \receiver can have different interpretations of the same message (Figure~\ref{fig:smcr}B).  
For example, people may perceive politeness cues differently depending on their cultural background \cite{thomas_cross-cultural_1983,riley_understanding_1984}. 
In our second application scenario, the interlocutors' perceptions of politeness are misaligned, 
and we aim to suggest paraphrases that reduce the potential for misinterpretation.

To successfully produce such \context-sensitive paraphrases, we need to depart from existing style transfer methodology (see \citeauthor{li_review_2020-1}, \citeyear{li_review_2020-1} for a survey, and \citeauthor{madaan_politeness_2020}, \citeyear{madaan_politeness_2020} for politeness transfer in particular).
First, since we must account for arbitrary levels of misalignment,
we need {\it fine-grained} control over the target stylistic \degree, as opposed to binary switches (e.g., from impolite to polite).
Second, we need to determine 
the target stylistic \degree
at run time, in an \emph{ad hoc} fashion, rather than assuming predefined targets. 

To overcome these new technical challenges, we 
start from the intuition that the same \degree of politeness can be conveyed through different combinations of 
{\it pragmatic strategies} \cite{lakoff_logic_1973,brown_politeness:_1987}, with some being more appropriate to the given \context than others.
We consider a classic two-step approach (Section~\ref{sec:method}), separating {\it planning}---choosing a viable combination of strategies that can achieve a desired stylistic \degree in a particular \context---, from the step of {\it realization}---incorporating this plan into generation outputs.
For a given fine-grained target stylistic \degree {{ (i.e., the \degree intended by the sender)}}, we find the optimal strategy plan via Integer Linear Programming (\ilp).  
We then realize this plan using a modification of the `Delete-Retrieve-Generate' (\drg) paradigm \cite{li_delete_2018} that allows for strategy-level control in generation.

Our experimental results indicate that in both our application scenarios, our method can suggest paraphrases that narrow the 
potential 
gap between the intended and perceived politeness, and thus better preserve the sender's intentions.  
These results show that automated systems have the potential to help people better convey their intentions in new communication \context{s}, and encourage further work exploring the feasibility and implications of such 
communication assistance applications. 

To summarize, in this work, we motivate and formulate the task of \context-sensitive \preserving paraphrasing (Section~\ref{sec:task}). Focusing on the case of pragmatic intentions, we introduce a model for paraphrasing with fine-grained politeness control (Section~\ref{sec:method}). We evaluate our method in two realistic communication scenarios to demonstrate the feasibility of the approach (Section~\ref{sec:eval}).

\section{Further Related Work}
\label{sec:related}

\xhdr{Style transfer} There has been a wide range of efforts in using NLP techniques to generate 
 alternative expressions, leading to tasks such as text simplification (see \citeauthor{shardlow_survey_2014}, \citeyear{shardlow_survey_2014} for a survey), 
	or more generally, paraphrase generation \cite[inter alia]{meteer_strategies_1988,quirk_monolingual_2004,fu_paraphrase_2019}.
When such paraphrasing effort is focused on the stylistic aspect, 
it is also referred to as text style transfer, which has attracted a lot of attention in recent years \cite[inter alia]{xu_paraphrasing_2012,ficler_controlling_2017,fu_style_2018,prabhumoye_style_2018}.
While these tasks are focused on satisfying specific
  predefined
  linguistic properties at the utterance-level, they are not designed for fine-grained adjustments to changing non-textual communication \context{s}.

\xhdr{Controllable generation} Style transfer or paraphrasing can both be seen as a special case of the broader task of
{\it controllable} text generation 
 \cite[inter alia]{hu_toward_2017,keskar_ctrl:_2019, dathathri_plug_2020}. While not focused on paraphrasing, relevant work in this area aims at controling the \degree of politeness for translation \cite{sennrich_controlling_2016} or dialog response \cite{niu_polite_2018}.

\xhdr{AI-assisted communications or writing} Beyond paraphrasing, AI tools have 
 been used to provide communication or writing assistance in diverse settings: from the mundane task of grammar and spell checking \cite{napoles_enabling_2019,stevens_grammar_2019}, to creative writing \cite{clark_creative_2018}, to negotiations
 \cite{zhou_dynamic_2019}, and has led to discussions of ethical implications~\cite{hancock_ai-mediated_2020-1}.

\xhdr{Models of communication} While Berlo's model provides the right level of abstraction for inspiring our application scenarios, many other models exist \cite{velentzas_communication_2014,barnlund_transactional_2017}, most of which are under the influence of the Shannon–Weaver model \cite{shannon_mathematical_1963}.

\section{Task Formulation}
\label{sec:task}

Given a {message} that a \texttt{\speaker} attempts to communicate to a \texttt{\listener} over a particular communication \texttt{\channel}, the task of \context-sensitive 
intention-preserving 
paraphrasing 
 is to generate a paraphrase that is more likely to convey the 
  intention of the \sender to the \receiver after transmission, under the 
  given 
  communication circumstance.

\xhdr{Formulation} To make this task more tractable, our formulation considers a single gradable stylistic aspect of the message that can be realized through a collection of \emph{pragmatic strategies} (denoted as $\strset$). 
While in this work we focus on politeness, other gradable stylistic aspects might include formality, humor
and certainty.

We can then formalize the relevant features of the communication \context as follows:
%
%\vspace{0.05 in}

\begin{enumerate}
	\item

	  For the communication \channel, we consider whether it can safely transmit each strategy $s \in \strset$. In particular, \cfunc$(s) = 1$ indicates that strategy $s$ is {\it safe} to use, whereas \cfunc$(s) = 0$ implies that $s$ is {\it at-risk} of being lost.

	\item For the \sender and \receiver, we quantify the 
	level of the stylistic aspect each of them perceive in a combination of pragmatic strategies via two mappings \fsend$:\mathcal{P}(\strset) \rightarrow \mathbb{R}$ and \mbox{\frec$:\mathcal{P}(\strset) \rightarrow \mathbb{R}$}, respectively, with $\mathcal{P}(\strset)$ denoting the powerset of $\strset$.
	
	\end{enumerate}

\vspace{0.05 in}

With our focus on politeness, our task can then be more precisely stated as follows: given an input message $m$, we aim to generate a politeness paraphrase for $m$, under the \context specified by (\fsend, \cfunc, \frec), such that the level of politeness perceived by the \listener is as close to that intended by the \speaker as possible.

We show that our theoretically-grounded formulation can model naturally-occurring challenges in communication,
by considering two possible application scenarios, each corresponding to a source of misalignment
 highlighted in Figure~\ref{fig:smcr}.

\xhdr{Application A: translated communication} We first consider the case of conversations mediated by translation services, where \channel-induced misunderstandings can occur (Figure \ref{fig:smcr}A): MT models may systematically drop certain politeness cues due to technical limitations or mismatches between the source and target languages. 

For instance, despite the difference in intended politeness level (indicated in parentheses) of the following two versions of the same request,\footnote{Annotations from 5 native speakers on a 7-point Likert scale ranging from \textsc{very impolite} to \textsc{very polite}.}   

\vspace{0.05 in}

\begin{tabular}{l}
     {\small Could you please proofread this article?} (\textsc{\small polite})\\
     {\small Can you proofread this article?} (\textsc{\small somewhat polite}) \\
\end{tabular}

\vspace{0.05 in}
\noindent Microsoft Bing Translator would translate both versions to the same utterance in Chinese.\footnote{Translating on May, 2020 to \begin{CJK*}{UTF8}{gbsn} 你能校对这篇文章吗？\end{CJK*}} By dropping the politeness marker `please', and not making any distinction between `could you' and `can you', the message presented to the Chinese receiver is likely to be more imposing than originally desired by the English sender.

To avoid such \channel-induced misunderstandings, the sender may consider using only strategies that are known to be safe with the specific MT system they use.\footnote{E.g., they might consider expressing gratitude (e.g., `thanks!') rather than relying on subjunctive (`could you').} However, since the inner mechanics of such systems are often opaque (and in constant flux), the sender would benefit from automatic guidance in constructing such paraphrases.

\xhdr{Application B: misaligned perceptions} We then consider the case when \senders and \receivers with differing perceptions interact.
Human perceptions of pragmatic devices are subjective, 
 and it is not uncommon to observe different interpretations of the same utterance, or pragmatic cues within, leading to misunderstandings \cite{thomas_cross-cultural_1983,kasper_cross-cultural_1990} (Figure~\ref{fig:smcr}B). 
For instance, a study comparing Japanese speakers'
 and American native English speakers'
 perceptions of English requests find that while the latter group takes the request `\textsl{May I borrow a pen?}' as strongly polite, their Japanese counterparts regard the expression as almost neutral \cite{matsuura_japanese_1998}.
In this case, if a native speaker still wishes to convey their good will, 
they need to find a paraphrase that would be perceived as strongly polite by Japanese speakers.

\section{Method}
\label{sec:method}

When compared to style transfer tasks, our \context-sensitive \preserving paraphrasing task gives rise to important new technical challenges. 
First, in order to minimize the gap in perceptions, we need to have fine-grained control over the stylistic aspect, as opposed to switching between two pre-defined binarized targets (e.g., polite vs. impolite).
Second, the desired degree of change is only determined at run-time,  
depending on the speaker's intention and on the communication \context.
We address these challenges by developing a method that allows for \emph{ad hoc} and \emph{fine-grained} paraphrase planning and realization.

Our solution starts from a strategy-centered view: instead of aiming for monolithic {\it style labels}, we think of {\it pragmatic strategies} as (stylistic) LEGO bricks.  
These can be stacked together in various combinations to achieve 
similar stylistic levels.
Depending on the \context, some bricks might,
or might not, be available. 
Therefore, given a message with an intended stylistic level, our goal is to find the optimal collection of 
available
bricks that can convey the same level---ad hoc
 fine-grained planning.
Given this optimal collection, we need to 
assemble it 
 with the rest of the message into a valid paraphrase---fine-grained realization.

\xhdr{Politeness strategies} In the case of politeness, we derive the set of pragmatic strategies from prior work \cite{danescu-niculescu-mizil_computational_2013,voigt_language_2017,yeomans_politeness_2019}. 
We focus on strategies that are realized through local linguistic \emph{markers}. 
For instance, the {\small \could} strategy can be realized through the use of markers like {\it could you} or {\it would you}.
In line with prior work, we further assume that markers realizing the same strategy has comparable {\it strength} in exhibiting politeness and are subject to the same constraints.
The full list of 18 strategies we consider (along with their example usages) can be found in Table \ref{tab:politeness_strategies_short}. Strategy extraction code is available in ConvoKit.\footnote{\url{https://convokit.cornell.edu}.}

\begin{table}[ht!]
\resizebox{\columnwidth}{!}{
\begin{tabular}{l l}
\bf{\small Strategy} &  \bf{\small Example usage} \\
    \toprule
    \actually & it {\bf actually} needs to be ... \\
    \just & i {\bf just} noticed that ... \\
    \affirmation & {\bf excellent point}, i have added it ...\\
    \apology& {\bf sorry} to be off-topic, but ... \\ 
    \btw & okay - {\bf by the way}, do you want me ...? \\ 
    \conj & {\bf so} where is the article ?\\
    \filler & {\bf uh}, hey, can you...?\\
    \forme &  is it alright {\bf for me} to archive it now?\\
    \foryou & i can fetch one {\bf for you} in a moment! ... \\
    \thx & {\bf thanks} for the info , ...\\
    \hello & {\bf hey} simon , help is needed if possible ... \\
    \hedges & {\bf maybe} some kind of citation is needed ... \\
    \can & {\bf can you} create one for me?\\
    \pls & can you {\bf please} check it?\\
    \plsstart & {\bf please} stop . if you continue ... \\
    \reassurance &  {\bf no problem}, happy editing. ...\\
    \could &  ..., {\bf could you} check? \\
    \swearing &  what {\bf the heck} are you talking about? \\
     \bottomrule
\end{tabular}}
    \caption{Politeness strategies we consider, along with example usage and example markers (in bold). More details for the strategies can be found in Table \ref{tab:politeness_strategies} in the Appendix.}
    \label{tab:politeness_strategies_short}
\end{table}

\vspace{0.1 in}

\xhdr{Ad hoc fine-grained planning}
Our goal is to find a target strategy combination that is estimated to provide a comparable pragmatic force to the 
sender's intention, using only strategies 
 appropriate in the current \context.
 To this end, we devise an Integer Linear Programming (\ilp) formulation that can efficiently search for the desired strategy combination to use (Section~\ref{sec:ilp}).

\xhdr{Fine-grained realization} 
To train a model that learns to merge the strategy plan into the 
original message
 in the absence of parallel data, we take inspirations from the 
\drg
  paradigm \cite{li_delete_2018}, originally proposed for style transfer tasks.
We adapt this paradigm to allow for direct integration with strategy-level planning, providing finer-grained control over realization (Section \ref{sec:generaion}).

\subsection{Fine-Grained Strategy Planning}
\label{sec:ilp}

Formally, given a message $m$ using a set of strategies $\strin$, under a \context specified by (\fsend, \cfunc, \frec), 
the planning goal is to find 
the set of strategies $\strout \subseteq \strset$ such that \cfunc$(s) = 1, \forall s \in \strout$  
---i.e., they can be safely transmitted through the communication \channel---and \fsend$(\strin) \approx$ \frec$(\strout)$---i.e., the resultant \receiver perception is similar to the intention the \sender had when crafting the original message. 

Throughout, we assume that both \interpretation mappings \fsend and \frec are linear functions:
\[
f_{send}(\strin) = \textstyle \sum_{s \in \mathcal{S}} a_s \indicator_{\strin} (s) + a_0
\]
\[
f_{rec}(\strout)= \textstyle \sum_{s \in \mathcal{S}} b_s \indicator_{\strout} (s) + b_0
\]

\noindent where the linear coefficients $a_s$ and $b_s$ are reflective of the strength of a strategy, as perceived by the \sender and \receiver, respectively.\footnote{We acknowledge that considering only linear models may result in sub-optimal estimations.}

\xhdr{Naive approach} One greedy type of approach to this problem is to consider each at-risk strategy $s \in \strin$
 at a time, and replace $s$ with a safe strategy $s'$ that is closest in strength. Mathematically, this can be written as $s' = \argmin_{\hat{s} \in \strset, f_c(\hat{s}) = 1} |a_s - b_{\hat{s}}|$.
 \noindent In our analogy, this amounts to reconstructing a LEGO model by replacing 
 each
  `lost' brick with the most similar brick that is available.

\xhdr{Our approach: \ilp formulation} The greedy approach, while easy to implement, can not consider solutions that involve an alternative \emph{combination} of strategies. 
In order to more thoroughly search for an appropriate strategy plan in the space of possible solutions in a flexible and efficient manner, we translate this problem into an \ilp formulation.\footnote{A brute force alternative would inevitably be less scalable.}
Our objective is to find a set of safe
 strategies $\strout$ that will be perceived by the \receiver as close as possible to the \sender's intention, i.e., one that that minimizes $|f_{send}(\strin) - f_{rec}(\strout)|$. 

To this end, we introduce a binary variable $x_s$ for each strategy $s$ in $\strset$, where we take $x_s = 1$ to mean that strategy $s$ should be selected to be present in the suggested alternative strategy combination $\strout$.  
We can identify the optimal value
of $x_s$ (and thus the optimal strategy set $\strout$) by solving the following \ilp problem:\footnote{All summations are over the entire strategy set $\mathcal{S}$. Throughout, we use the \texttt{PuLP} package \cite{mitchell_pulp_2011} with \texttt{GLPK} solver to obtain solutions.}

\vspace{0.05 in}

{\centering
	\resizebox{1.03\columnwidth}{!}{
	\begin{tabular}{l l}
	\textsc{\small min} & \textnormal{\hspace{0.15 in}} $y$  \vspace{0.1 in} \\
	{\small subj to} & $\textstyle (\sum a_s \indicator_{\strin} (s) + a_0) - (\sum b_s x_s + b_0) \leq y $ \vspace{0.05 in}\\
		& $\textstyle (\sum b_s x_s + b_0) - \textstyle (\sum a_s \indicator_{\strin} (s) + a_0) \leq y $ \vspace{0.05 in}\\
		& $x_s \leq f_c (s), x_s \in \{0, 1\}, \forall s \in \mathcal{S} $
	\end{tabular}}
}
\vspace{0.05 in}

\noindent which is a rewriting our objective to minimize $|f_{send}(\strin) - f_{rec}(\strout)|$ that satisfies the linearity requirement of \ilp via an auxiliary variable $y$, and where our target variables $x_s$ 
replace the indicator function $\indicator_{\strout} (s)$ in the linear expression of $f_{rec}$.

The \channel constraints are encoded by the additional 
 constraints $x_s \leq f_c (s)$, allowing only safe strategies (i.e., those for which $f_c (s)=1$) to be included. 
Additional strategy-level constraints can be similarly specified through this mechanism to obtain strategy plans that are easier to realize in natural language (Section \ref{sec:constraints} in the Appendix).

\begin{figure*}
 \centering
	\includegraphics[width=\textwidth]{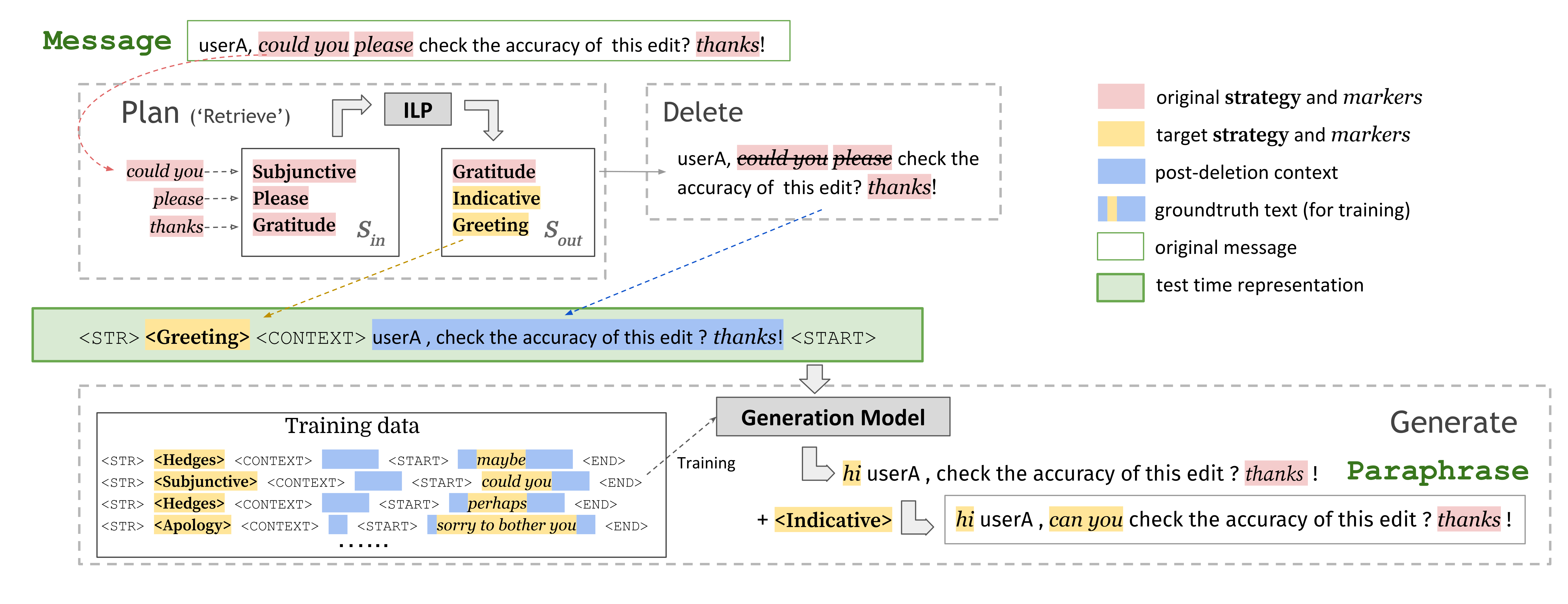}
\caption{Sketch of our pipeline for generating politeness paraphrases. Given an 
input message,
 we first identify the politeness \setlength{\fboxsep}{0pt}\colorbox{lightpink}{strategies} ($\strin$) and the corresponding \setlength{\fboxsep}{0pt}\colorbox{lightpink}{markers} it contains. In the {\bf plan} step, we use \ilp to compute a target {strategy} combination ($\strout$) that is appropriate 
under the \context. We then {\bf delete} markers corresponding to strategies that need to be removed to obtain the \setlength{\fboxsep}{0pt}\colorbox{lightblue}{post-deletion \content}. Finally, we sequentially insert the \colorbox{lightyellow}{new strategies} from the \ilp solution into this context 
to {\bf generate} the {final output}.}
 \label{fig-methodology}
\end{figure*}

\subsection{Fine-Grained Realization}
\label{sec:generaion}

To transform the \ilp solutions into natural language paraphrases, we build on the general \drg
framework, which has shown strong performance in style transfer without 
 parallel data.\footnote{Since the politeness strategies we consider are local, they fit the assumptions of \drg framework well.} 
We modify this framework to allow for the fine-grained control 
needed to realize strategy plans.

As the name suggests, the vanilla \drg framework consists of three steps. With {\it delete}, 
lexical \attr (n-grams) that are strongly indicative of style are removed, resulting in a `style-less' intermediate text. 
In the {\it retrieve} step, target \attr are obtained by considering 
those
 used in training examples that are similar to the input but exhibit the desired property (e.g., target sentiment valence). 
Finally, in the {\it generate} step, the generation model merges the desired target \attr with the style-less intermediate text to create the final output.

Importantly, the \drg framework is primarily designed to select to-be-inserted markers
based on pre-defined binary style classes. As such, it cannot directly allow the
ad hoc
 fine-grained control needed by our application. 
We now explain our modifications in detail (follow the sketch of our pipeline in Figure~\ref{fig-methodology}):

\xhdr{Plan (instead of Retrieve)} 
We first perform a Plan step, which substitutes the Retrieve step in \drg, but it is performed first in our pipeline as our version of the Delete step is dependent on the planning results. For an input message, we identify the politeness strategies it contains and set up the corresponding \ilp problem (Section \ref{sec:ilp}) to obtain their {\it functional alternatives}. By factoring in the communication \context into the \ilp formulation, we obtain an ad hoc strategy plan to achieve the intended level of politeness. This is in contrast with the Retrieve step in DRG, in which target markers from similar-looking texts are used for direct {\it lexical substitution}.

\xhdr{Delete} Instead of identifying style-bearing lexical \attr  
to delete 
with either frequency-based heuristics \cite{li_delete_2018}, or sentence context \cite{sudhakar_transforming_2019}, we rely on linguistically informed politeness strategies.
To prepare the input message for the new strategy plan, we compare the strategy combination from the \ilp solution with those originally used. 
We then {\it selectively} remove strategies that do not appear in the \ilp solution by deleting the corresponding markers found in the input message. As such, in contrast with \drg, our post-deletion \content is not necessarily style-less, and it is also
possible that no deletion is performed. 

\xhdr{Generate} Finally, we need to generate fluent utterances that integrate the strategies identified by the Plan step into the post-deletion \content.
To this end, we adapt \gst \cite{sudhakar_transforming_2019},
whose generation model is fine-tuned
to learn to integrate lexical markers into post-deletion \content.  
To allow smooth integration of the \ilp solution, we instead train the generation model to incorporate politeness strategies directly. %

Concretely, training data 
exemplifies how each target strategy can be integrated into various post-deletion \content{s}. 
This data is constructed by finding \textsc{\small groundtruth} utterances containing markers corresponding to a certain \textsc{\small strategy}, and removing them to obtain the post-deletion \textsc{\small \content}.
These training instances are represented as \textsc{\small (strategy, \content, groundtruth)} tuples separated by special tokens 
(examples in Figure \ref{fig-methodology}).
The model is trained to minimize the reconstruction loss.\footnote{We adapted the implementation from \citet{sudhakar_transforming_2019} to incorporate our modification described above, and we use their default training setup.}

At test time, we sequentially use the model to integrate each \textsc{\small strategy} from the plan into the post-deletion \textsc{\small \content}.
We perform beam search of size 3 for each strategy we attempt to insert and select the output that best matches the intended \degree of politeness as the paraphrase suggestion.\footnote{We set an upper bound of at most 3 new strategies to be introduced to keep sequential insertion computationally manageable. This is a reasonable assumption for short utterances.}

\section{Evaluation}
\label{sec:eval}

To test the feasibility of our approach, we set up two { parallel} experiments {{with different circumstance specifications, so that each illustrates}}
  one potential source of misalignment as described in Section~\ref{sec:task}.\footnote{Code and data is available at \url{https://github.com/CornellNLP/politeness-paraphrase}.}

\subsection{Experiments}
\label{sec:exp}

\xhdr{Data} We use the annotations from the Wikipedia section of the Stanford Politeness Corpus (henceforth annotations) to train \interpretation models that 
will
 serve as 
approximations of 
 \fsend and \frec.
 In this corpus, each utterance was rated by 5 annotators on a 25-point scale from very impolite to very polite, which we rescale to the $[-3,3]$ range.

To train the generation model, we randomly sample another (unannotated) collection of talk-page messages from WikiConv \cite{hua_wikiconv_2018}. 
For each strategy, we 
use
 1,500 disjoint instances for training 
		(27,000 in total, 2000 used for validation)
and 
additionally resource 200 instances per strategy 
 as test data. Both the Stanford Politeness Corpus and WikiConv are retrieved from ConvoKit \cite{chang_convokit_2020}.

\xhdr{Experiment A: translated communication} We first consider MT-mediated English to Chinese communication using Microsoft Translator, where \channel-induced misunderstandings may occur.

For this specific \channel, we estimate its \cfunc by performing back-translation\footnote{{{Back-translation refers to the process of translating the translated text back into the source language.}}} \cite{tyupa_theoretical_2011}
on a sampled set of utterances from the collection of Stack Exchange requests from the Stanford Politeness Corpus. We consider a strategy $s$ to be at-risk under this MT-mediated \channel 
if the majority of messages using $s$ have back-translations that no longer uses it.
We identify four at-risk strategies, leading to the following channel specification: \cfunc$(s) = 0$, if $s \in $ \{{\small \could, \pls, \filler, \swearing}\}; \cfunc$(s) = 1$ otherwise.  

For the \sender and the \receiver, we make the simplifying assumption that they both perceive politeness similar to a prototypical `average person' (an assumption we address in the next experiment), and take the average scores from the annotations to train a linear regression model $f_{avg}$ to represent the \interpretation model, i.e., \fsend = \frec =  $f_{avg}$.

We retrieve test data corresponding to the four at-risk strategy types as test messages ($4 \times 200$ in total). 
We estimate the default perception gap
(i.e., when \textbf{no intervention} takes place) by comparing the intended \degree of politeness in the original message and the \degree of politeness of its back-translation, which roughly approximates what the \listener sees after translation,
following \citet{tyupa_theoretical_2011}.
 This way, we can avoid having to compare
  politeness levels across different
  languages.
\begin{table*}[t!]
\centering
\resizebox{2\columnwidth}{!}{
\begin{tabular}{l  c c c c | c c c c}
\toprule 
  & \multicolumn{4}{c}{\bf Translated communication (A)} & \multicolumn{4}{c}{\bf Misaligned perceptions (B)} \\
  & {\bf \mae$_{plan}$} &  {\bf \mae$_{gen}$} & {\small BLEU-s} & {\small \textsc{\#-added}} & {\bf \mae$_{plan}$} & {\bf \mae$_{gen}$} & {\small BLEU-s} & {\small \textsc{\#-added}}  \\
  \midrule
  No intervention & 0.43 & 0.43 & 64.2 & 0 & 1.01 & 1.01 & 100 & 0 \\
  \midrule
  Retrieval (\drg) & 0.66 & 0.61 & 74.7 & 1.09 &  0.81 & 0.76 & 72.0 & 1.07 \\
  Greedy & 0.35 & 0.35 & 73.5 & 1.20 & 0.48 & 0.47 & 70.3 & 1.82 \\
  \ilp-based & {\bf 0.14} & {\bf 0.21} & 67.0 & 2.38 & {\bf 0.03} & {\bf 0.12} & 68.8 & 2.30 \\
\bottomrule
\end{tabular}}
\caption{Our method is the most efficient at reducing the potential for misalignment (bolded, t-test $p < 0.001$).}

    \label{tab:generation_eval_large}

\end{table*}

\begin{table*}[ht!]
\centering
\resizebox{2\columnwidth}{!}{
\begin{tabular}{l l l}

       & {\bf Input / Output} & {\bf Gap} \vspace{0.05 in} \\
  
  \toprule
     \bf{\small Experiment A} 

                      & \setlength{\fboxsep}{0pt}\colorbox{lightpink}{could you} clarify what type of image is requested of centennial olympic park? \colorbox{lightpink}{thanks}! & 0.23 \\ 

                      & \setlength{\fboxsep}{0pt}\colorbox{lightyellow}{can you{\color{lightyellow}{h}}}clarify what type of image is requested of centennial olympic park \colorbox{lightyellow}{for me} ? \colorbox{lightpink}{thanks} ! & 0.11 \vspace{0.1 in} \\

                    & where \setlength{\fboxsep}{0pt}\fcolorbox{white}{lightpink}{the hell} did i say that? i was referring to the term `master'. & 1.30 \\

                    &   \setlength{\fboxsep}{0pt}\fcolorbox{white}{lightyellow}{so{\color{lightyellow}{h}}}where did i \fcolorbox{white}{lightyellow}{actually} say that ? i was referring to the term `master'. & 0.70 \vspace{0.1 in} \\

     \midrule

  \bf{\small Experiment B} & \setlength{\fboxsep}{1pt}\fcolorbox{white}{lightpink}{thanks{\color{lightpink}{p}}} for accepting. how and when do we start? \fcolorbox{white}{lightpink}{sorry} for the late reply. & 1.30 \\

              & \setlength{\fboxsep}{0pt}\fcolorbox{white}{lightyellow}{hi, no problem}. \fcolorbox{white}{lightpink}{thanks{\color{lightpink}{p}}} for accepting. how and when do we start? & 0.03 \vspace{0.1 in} \\

              & i'd like to try out kissle, so \setlength{\fboxsep}{0pt}\colorbox{lightpink}{would you} \colorbox{lightpink}{please} add me to [it]? \colorbox{lightpink}{thanks}. & 1.06 \\

              & \setlength{\fboxsep}{0pt}\colorbox{lightyellow}{hi !\color{lightyellow}{g}} i ' d like to try out kissle , so \colorbox{lightyellow}{{\color{lightyellow}{h}}will you just} add me to [it]? & 0.01 \vspace{0.1 in} \\

     \midrule

 \bf{\small Error case} & \setlength{\fboxsep}{0pt}\colorbox{lightpink}{hi, would you please} reply to me at the article talk page? \colorbox{lightpink}{thanks}. & 0.97 \\

                  &  \setlength{\fboxsep}{0pt}\colorbox{lightyellow}{good idea . sorry} , \colorbox{lightpink}{would you please} reply to me at the article talk page \colorbox{lightyellow}{for you} ? & 0.01 \vspace{0.05 in} \\

     \bottomrule
    \end{tabular}}
    \caption{Example generation outputs (we highlight the \setlength{\fboxsep}{0pt}\fcolorbox{white}{lightpink}{original} and \setlength{\fboxsep}{0pt}\fcolorbox{white}{lightyellow}{newly introduced} markers through which the strategies are realized). For reference, we also show the (estimated) gap between the \sender's intention and the \receiver's perception after transmission.  
    More example outputs and error cases are shown in Tables~\ref{tab:generation_additional} and \ref{tab:generation_errors} in the Appendix.} 
    \label{tab:generation_outputs}
\end{table*}

\xhdr{Experiment B: misaligned perceptions} We then consider communication between individuals with misaligned politeness perceptions.
Under this circumstance, we assume a perfect \channel, which allows any strategy to be safely transmitted, i.e., \cfunc$(s) = 1, \forall s \in S$. 
We then consider the top 5 most prolific annotators
as potential 
\senders and \receivers.
 To obtain \fsend (and \frec), we use the respective annotator's  
 annotations 
 to train an 
individual linear regression model.\footnote{Details about the choice of annotators and their perception models are described in Section \ref{sec:annotators} in the Appendix. 
{ While in practice individual \interpretation models may not be available, they could potentially be approximated based on annotations from people with similar (cultural) backgrounds.}}

We take all permutations of \texttt{(\speaker, \listener)} among the chosen annotators, resulting in 20 different directed pairs. For each 
pair,
we select as test data the top 100 utterances with the greatest (expected) perception gap in the test set.
We take the default perception gap within the pair (with \textbf{no intervention}) as the difference between the \sender's intended \degree of politeness (as judged by \fsend) and the \receiver's perceived \degree of politeness (as judged by \frec).

\xhdr{Baselines} Beyond the base case with no intervention, we consider baselines with different degrees of planning. We first consider binary-level planning by
directly applying vanilla
 \textbf{\drg} in our setting: for each message, we retrieve from the generation training data the most 
  similar utterance that has the same politeness polarity as the input message,\footnote{We use $f_{avg}$ to determine the binary politeness polarity.} and take the strategy combination used within as the new strategy plan. 
  We then consider a finer-grained strategy planning based on the naive \textbf{greedy} search, for which we substitute each at-risk strategy by an alternative that is the closest in strength. To make fair comparisons among different planning approaches, we apply the same set of 
  constraints (either \context-induced or generation-related) we use with \ilp.\footnote{We note that even if we do not enforce these additional constraints, the baselines still under-perform the \ilp solution.}

\xhdr{Evaluation} We compare the paraphrasing outputs 
using
 both automatic and human evaluations.   
First, we consider our {\it {main objective}}: how effective each model is at reducing the 
potential
gap between intended and perceived politeness. 
We compare the predicted perceived politeness levels of paraphrases generated by each model with the intended politeness levels of the original inputs in terms of mean absolute error ({\bf \mae$_{gen}$}), with smaller values corresponding to smaller gaps.  
We additionally evaluate the (pre-generation) quality of the planned strategy set ({\bf \mae$_{plan}$}) to account for cases in which the plan is not perfectly realized.

To check the extent to which the generated paraphrases could be readily used, we assess how natural they sound to humans.
We sample 100 instances from each set of the generated outputs and ask one non-author native English speaker to judge their naturalness
on a scale of 1 to 5 (5 is very natural). 
	The task is split among two annotators, and we obtain one annotation for each utterance. Each annotator was presented with an even distribution of retrieval-based, greedy-based and \ilp-based generation outputs, and was not given any information on how the outputs are obtained.\footnote{Table \ref{tab:generation_instruction} in the Appendix shows the exact instructions.}

To validate that the original content is not drastically altered, we report \textsc{bleu} scores \cite{papineni_bleu_2002} obtained by comparing the generation outputs with the original message (\textsc{bleu}-s), 
Additionally, we provide a rough measure of how `ambitious' the paraphrasing plan is by counting the number of new strategies that are \textsc{added}.

\xhdr{Results} Table \ref{tab:generation_eval_large} shows that our \textbf{\ilp-based} method is capable of significantly reducing the potential gap in politeness \interpretation{s} between the \speaker and the \listener, 
in both experiments (t-test $p < 0.001$).  
The comparison with the baselines underlines the virtues of supporting fine-grained planning: 
the effectiveness of the eventual paraphrase is largely determined by the quality of the strategy plan. 
This can be seen by comparing across the \mae$_{plan}$ column which shows misalignments that would result if the plans were perfectly realized.
Furthermore, when planning is done { too coarsely}
 (e.g., at a binary granularity for vanilla \drg),  
 the { resultant misalignment} 
  can be even worse than not intervening at all (for translated communication). 

At the same time, the paraphrases remain mostly natural, with the average annotator ratings generally fall onto 
`mostly natural'
 category for all generation models. 
The exact average ratings are 4.5, 4.2, and 4.2 for the retrieval-based, greedy-based, and \ilp-based generation respectively.  
These generation outputs also largely preserve the content of the original message, as indicated by the relatively high \textsc{bleu}-s scores.\footnote{As a comparison point, we note that the outputs of all methods have higher \textsc{bleu}-s scores than the back-translations. We have also verified that the generated paraphrases preserve more than 90\% of the non-marker tokens, further suggesting the degree of content preservation.}
Considering that the \ilp-based method (justifiably) implements a
more ambitious plan than the baselines (compare \textsc{\small \#-added}), it is expected to depart more from the original input; in spite of this, the difference in naturalness is small.

\subsection{Error Analysis}
\label{sec:analysis}

By inspecting the output (examples in Tables \ref{tab:generation_outputs}, \ref{tab:generation_additional} and \ref{tab:generation_errors}), we identify a few issues that are preventing the model to produce ideal 
 paraphrases, opening avenues for future improvements: 

\xhdr{Available strategies} Between the two experimental conditions reported in Table~\ref{tab:generation_eval_large}, we notice that the performance ({\bf \mae$_{gen}$})
 is worse for the case of translated communication. 
A closer analysis reveals that this is mostly due to a particularly hard-to-replace at-risk strategy, \swearing, which is one of the few available strategies that have
strong negative politeness valence. 
The strategy set we operationalize 
 is by no means exhaustive.
 Future work can consider a more comprehensive set of strategies, or even individualized collections, to allow more diverse expressions. 

\xhdr{Capability of the generation model} From a cursory inspection,
 we find that the generation model has learned to incorporate the planned strategies, either by realizing simple maneuvers such as \textit{appending} markers at sentence boundaries, to the more complex actions such as {\it inserting} relevant markers in reasonable positions within the messages (both exemplified in Table~\ref{tab:generation_outputs}). However, 
 the generation model does not always fully execute the strategy plan, and can make inappropriate insertions, especially in the case of the more ambitious \ilp solutions.
We anticipate more advanced generation models may help further improve the quality and naturalness of the paraphrases. 
Alternatively, dynamically integrating the limitations of the generation model as explicit planning constraints might lead to solutions that are easier to realize.

%

% We show examples of typical errors, as well as suboptimal outputs in Table  in the Appendix. 

\section{Discussion}
\label{sec:discussion}

In this work, we motivate and formulate the task of \context-sensitive intention-preserving paraphrasing and develop a methodology that shows promise in helping people more accurately communicate politeness
under different communication settings. The results and limitations of our method 
open up several natural directions for future work.

\xhdr{Modeling politeness perceptions} We use a simple linear regression model to approximate how people internally interpret politeness and restrict our attention to only the set of local politeness strategies. 
  Future work may consider more comprehensive modeling of how people form politeness perceptions or obtain more reliable causal estimates for strategy strength \cite{wang_when_2019}.

\xhdr{Task formulation} We make 
several simplifying assumptions in our task formulation. 
First, we focus exclusively on
 a gradable stylistic aspect that is mostly decoupled from the content \cite{kang_xslue_2019}, reducing the complexity required from both the \interpretation and the generation models. 
Future work may
 consider more complex stylistic aspects and strategies that are more tied to the content, such as switching from active to passive voice. 
  Second, we consider 
   binary channel constraints, but in reality, the channel behavior is often less clear-cut.
   Future work can aim to propose more general 
    formulations that encapsulate more properties of the circumstance.

	\xhdr{Forms of assistance} While we have focused on offering paraphrasing options as the form of assistance, it is not the only type of assistance possible. As our generation model may not (yet) match the quality of human rewrites, there can be a potential trade-off. While an entirely automatic assistance option may put the least cognitive load on the user, it may not produce the most natural and effective rewrite, which may be possible if humans are more involved.
    Hence, while we work towards providing fully automated suggestions, we might also want to utilize the language ability humans possess and consider assistance approaches in the form of interpretable (partial) suggestions.

\xhdr{Evaluation} In our experiments, we have relied exclusively on model predictions to estimate the level of misalignment in politeness perceptions. Given the fine-grained and individualized nature of the task, using humans to ascertain the politeness of the outputs would require an extensive and relatively complex annotation setup (e.g., collecting fine-grained labels from annotators with known backgrounds for training and evaluating individualized perception models). Furthermore, to move towards more practical applications, we would also need to conduct communication-based evaluation \cite{newman_communication-based_2020} in addition to annotating individual utterances.
  Future work can consider adapting experiment designs from prior work \cite{gao_two_2015,hohenstein_ai-supported_2018} 
 to establish the impact of offering such intention-preserving paraphrases in real conversations, potentially by considering downstream outcomes.

\xhdr{Bridging the gaps in perceptions} While we focus on politeness strategies,
they are not the only \context-sensitive linguistic signals that may be lost or altered during transmission, nor the only type 
that are subject to individual or cultural-specific \interpretation{s}. Other examples commonly observed in communication include, but are not limited to, formality \cite{rao_dear_2018} and emotional tones \cite{chhaya_frustrated_2018,raji_what_2020}. As we are provided with more opportunities to interact with people across 
cultural and language barriers, the risk of misunderstandings in communication also grows \cite{chang_dont_2020-1}. Thus, it is all the more important to develop tools to mitigate such risk and help foster mutual understandings.

\vspace{0.2 in}

\xhdr{Acknowledgments} We thank Jonathan P. Chang, Caleb Chiam, Hajin Lim, Justine Zhang, and the reviewers for their helpful comments, Kim Faughnan and Ru Zhao for the naturalness annotations. We also thank Sasha Draghici for showing us his extensive LEGO collection, which was an inspiration for the analogy used in this paper. This research was supported in part by NSF CAREER award IIS-1750615 and NSF Grant IIS-1910147. 

\bibliographystyle{acl_natbib}
\bibliography{liye-translation}

\appendix
\section*{Appendices}
\renewcommand{\thetable}{A\arabic{table}}
\setcounter{table}{0}
\setcounter{section}{0}

\section{Politeness Strategies}

\begin{table*}[ht!]
\resizebox{2\columnwidth}{!}{
\begin{tabular}{l r l l l}
    \bf{\small Strategy} & {\bf \small Coeff.} & \bf{\small Example markers} & \bf{\small Delete mode} & \bf{\small Example usage}\\
    \toprule
    \actually & -0.358 & really, actually & token & it {\bf actually} needs to be ... \\
    \just & -0.004 & just & token & i {\bf just} noticed that ... \\
    \affirmation & 0.171 & ok, good [work] & segment & {\bf excellent point}, i have added it ...\\
    \apology & 0.429 & sorry, [i] apologize & segment & {\bf sorry} to be off-topic but ... \\ 
    \btw & 0.331 & by the way, btw & token & okay - {\bf btw}, do you want me ...? \\ 
    \conj & -0.245 & so, and, but & token &  {\bf so} where is the article ?\\
    \filler &  -0.245 & hmm, um &token & {\bf uh}, hey, can you...?\\
    \forme & 0.128 & for me &token & is it alright {\bf for me} to archive it now?\\
    \foryou & 0.197 & for you & token & i can fetch one {\bf for you} in a moment! ... \\
    \thx & 0.989 & thanks, [i] appreciate & segment & {\bf thanks} for the info , ...\\
    \hello & 0.491 & hi, hello &token & {\bf hey} simon , help is needed if possible ... \\
    \hedges & 0.131 & possibly, maybe &token & {\bf maybe} some kind of citation is needed ... \\
    \can & 0.221 & can you, will you & token & {\bf can you} create one for me?\\
    \pls & 0.230 & please & token & can you {\bf please} check it?\\
    \plsstart & -0.209 & please & token & {\bf please} stop . if you continue ... \\
    \reassurance & 0.668 & no worries & segment & {\bf no problem}, happy editing. ...\\
    \could & 0.454 & could you, would you & token & ..., {\bf could you} check? \\
    \swearing & -1.30 & the hell, fucking & token & what {\bf the heck} are you talking about? \\
    \bottomrule
\end{tabular}}

    \caption{Local politeness strategies being considered. For each strategy, we show its corresponding coefficients in the linear regression model, example markers, together with example usages.}
    \label{tab:politeness_strategies}
\end{table*}

We show the complete list of politeness strategies we use in Table \ref{tab:politeness_strategies}, together with the coefficients for the average model $f_{avg}$ used in Experiment A is shown in Table \ref{tab:politeness_strategies}. 

Recognizing that individual markers may not always fully encompass the politeness-bearing portion of the text, we consider two modes of deletion depending on strategy (Table~\ref{tab:politeness_strategies}): \textit{token} mode deletes only the identifier marker, whereas in \textit{segment} mode the whole sentence segment (as defined by within-sentence punctuations) will be removed:

\vspace{0.05 in}

\noindent \begin{tabular}{l l}
     
     \texttt{\bf \small Token mode} & {\small Can you \st{\bf please} explain?}\\
     \texttt{\bf \small Segment mode} & {\small \st{{\bf Thanks} for your help,} {\small I will try again.}} \\

\end{tabular}

\vspace{0.05 in}

\section{Prolific Annotators} 
\label{sec:annotators}

For experiment B, we sample the top five most prolific annotations from the Wikipedia section of the Stanford Politeness Corpus, with the most prolific one having annotated 2,063 instances, and the least prolific among the five having 715 annotations. 

When training individual \interpretation models, we note that some less frequently used strategies tend to be under annotated at the individual level, and may thus create artificially high difference in coefficients. We thus use the coefficient from the average model for any strategy that is annotated for less than 15 times by the individual annotator.

\section{Additional Details on \ilp}
\label{sec:constraints}

We consider a few linguistic constraints to help exclude some counter-intuitive strategy combinations. It should be noted that, with increased quality of a generation model, or by dynamically integrating the limitation of the generation model into the planning step, the process of inserting such additional constraints may be automated:

\xhdr{\bf Negativity constraint} While our simple linear model estimates the level of politeness by the aggregated effects of all strategies used regardless of their polarity, humans are known to have a negativity bias \cite{baumeister_bad_2001}: while the presence of polite markers in an otherwise impolite utterance may soften the tone, the use of a negative marker in an otherwise polite utterance may be overshadowing. As a result, when an input is judged to be positive in politeness, we consider the additional constraint to exclude use of negative strategies, i.e., $x_s = 0, \forall s \in \{s: b_s < 0\}$.

\xhdr{Subjunctive and Indicative constraint} Admittedly, among the set of markers we consider, some are more decoupled from contents than others---while removing {\it just} is almost guaranteed to keep the original meaning of the sentence intact, for an utterance that starts with either \could or \can, e.g., {\it could you clarify?}, simply removing {\it could you} would have already made its meaning ambiguous.\footnote{For instance, {\it can I clarify?} and {\it can you clarify?} would both be linguistically plausible requests containing {\it clarify}, yet they differ significantly in meaning.} To account for this, we add the constraint that the use of \could and \can should be substituted within themselves, i.e., $x_{\could} + x_{\can} = \indicator_{\strin(\could)} + \indicator_{\strin(\can)}$.\footnote{We acknowledge that under certain circumstances, this constraint may be impossible to fulfill.}

\section{Details on Human Evaluations}
 
To evaluate on the naturalness of the generated text, we ask two non-author native speaker for naturalness ratings on a scale of 1 (very unnatural) to 5 (very natural). 
	The exact instruction is shown in Table \ref{tab:generation_instruction}.

\begin{table}[ht!]
\begin{tabular}{p{7.2 cm}}
	\toprule
  	Ignoring punctuations, typos, and missing context, on a scale of 1-5, how natural does the text sound? \vspace{0.05 in} \\
		5. {\bf Very natural}: It’s possible to imagine a native speaker sending the message online.\\
		4. {\bf Mostly natural}: While there are some minor errors, simple edits can make it become `very natural'.\\
		3. {\bf Somewhere in between}: While the text is comprehensible, it takes more involved edits to make it sound natural.\\
		2. {\bf Mostly unnatural}: There are significant grammatical issues that make the text almost not comprehensible. \\
		1. {\bf Very unnatural}: Entirely broken English.\\
	\bottomrule
\end{tabular}

    \caption{Instruction for naturalness annotations.}
    \label{tab:generation_instruction}
\end{table}

\section{Additional Generation Examples}

We show additional generation outputs in Table~\ref{tab:generation_additional}, and a categorization of failure cases in Table~\ref{tab:generation_errors}.

\begin{table*}[ht!]
\resizebox{2\columnwidth}{!}{
\begin{tabular}{p{3.4 cm} p{10 cm} c}
  
  {\bf Strategy plan} & {\bf Input (upper) / Output (lower)} & {\bf Score} \\
    \toprule

  {\it \pls, \could, \thx} & \setlength{\fboxsep}{0pt}\colorbox{lightpink}{could you} then \colorbox{lightpink}{please} make some contributions in some of your many areas of expertise? \colorbox{lightpink}{thanks}. & \vspace{0.05 in} \\ 
  
  \underline{\hello}, \underline{\could}, \underline{\just}, \underline{\forme}  & \setlength{\fboxsep}{0pt}\colorbox{lightyellow}{hi\textcolor{lightyellow}{y}}, \colorbox{lightpink}{could you} then \colorbox{lightyellow}{just} make some contributions \colorbox{lightyellow}{for\textcolor{lightyellow}{y}me} in some of your many areas of expertise ?  & 5 \vspace{0.2 in} \\

  {\it \pls} & can someone \setlength{\fboxsep}{0pt}\colorbox{lightpink}{please} explain why there's a coi tag on this article? it's not evident from the talk page. & \vspace{0.05 in}  \\
 
   \underline{\forme}, \hedges & can someone explain why there ' s a coi tag on this article \colorbox{lightyellow}{for me}? it ' s not apparent from the talk page . & 5 \vspace{0.2 in} \\ 

  {\it \conj, \filler} & \setlength{\fboxsep}{1pt}\colorbox{lightpink}{uh}...ok...whatever...did you get that user name yet?\colorbox{lightpink}{\textcolor{lightpink}{h}or} do you prefer hiding behind your ip?  \vspace{0.05 in} \\
  \underline{\actually}, \btw, \underline{\conj}, \plsstart & ok . . . whatever . . . did you \setlength{\fboxsep}{1pt}\colorbox{lightyellow}{actually} get that user name yet ?\colorbox{lightpink}{\textcolor{lightpink}{h}or} do you prefer hiding behind your ip ? & 5 \vspace{0.2 in} \\

   {\it \pls, \could}  & \setlength{\fboxsep}{0pt}\colorbox{lightpink}{could you} \colorbox{lightpink}{please} stop your whining, and think about solutions instead? tx. \vspace{0.05 in} \\
  
  \underline{\btw}, \underline{\hedges}, \underline{\can} & \setlength{\fboxsep}{0pt}\colorbox{lightyellow}{btw\textcolor{lightyellow}{y}}, \colorbox{lightyellow}{can you maybe} stop your whining , and think about solutions instead ? tx . & 5 \vspace{0.2 in} \\

 	\bottomrule
\end{tabular}}
	\caption{Additional examples from the generation outputs, together with strategy information ({\it original strategy combination} for inputs in italics, \underline{realized strategies} underlined for outputs) and naturalness scores. We also highlight the \setlength{\fboxsep}{0pt}\fcolorbox{white}{lightpink}{original} and \setlength{\fboxsep}{0pt}\fcolorbox{white}{lightyellow}{newly introduced} markers through which the strategies are realized. Refer to Table~\ref{tab:generation_errors} for common types of failure cases.}
    \label{tab:generation_additional}
\end{table*}

\begin{table*}[ht!]
\resizebox{2\columnwidth}{!}{
\begin{tabular}{p{2.5cm} p{11.2Cm} c}
  
  {\bf Error type} & {\bf Input (upper) / Output (lower)} & {\bf Score}\\
    \toprule

  {\small \texttt{Grammatical}} & the bot seems to be down again. could you give it a nudge? &  \\ 
  {\small \texttt{mistake}}  & the bot seems to be down again . {\bf maybe} can you give it a nudge for me ? & 3 \vspace{0.1 in} \\ 

  & i see you blocked could you provide your rationale? thanks - () \\ 
  & i see you blocked {\bf provide} your rationale ? ( {\bf please} ) & 2 \\ 

  \midrule

  {\small \texttt{Strategy}} & hello, this image has no license info, could you please add it? thank you. &  \\ 

   {\small \texttt{misfit}} & hello , this image has no license info , {\bf sorry} . could you add it {\bf for you} ? thank you . & 3 \vspace{0.1 in} \\ 

   & can you please review this or block or have it reviewed at ani? thank you \\
   & {\bf no worries . sorry} , can you review this or block or have it reviewed for me at ani ? & 3 \\

  \bottomrule
\end{tabular}}
  \caption{Examples demonstrating two representative error types with naturalness scores. {\small \texttt{Grammatical mistake}} represents cases when the \attr are in inappropriate positions or introduce errors to the sentence structure. {\small \texttt{Strategy misfit}} represents cases when the use of suggested strategies (regardless of choice of \attr to realize them) do not seem appropriate. Problematic portions of the outputs are in bold.}
    \label{tab:generation_errors}
\end{table*}

\end{document}